%% file: main.tex
\pdfoutput=1
\documentclass{article}
\usepackage[final,nonatbib]{neurips_2024}
\usepackage[numbers]{natbib}

\usepackage[utf8]{inputenc} % allow utf-8 input
\usepackage[T1]{fontenc}    % use 8-bit T1 fonts
\usepackage{hyperref}       % hyperlinks
\usepackage{url}            % simple URL typesetting
\usepackage{booktabs}       % professional-quality tables
\usepackage{amsfonts}       % blackboard math symbols
\usepackage{nicefrac}       % compact symbols for 1/2, etc.
\usepackage{microtype}      % microtypography
\usepackage{xcolor}         % colors
\usepackage[pdftex]{graphicx}
\usepackage{amsmath}
\usepackage{amsfonts}
\usepackage{amssymb}
\usepackage{float}
\usepackage{caption}
\usepackage{color}
\usepackage{algorithmic}
\usepackage{algorithm}
\usepackage{multirow}
\usepackage{subcaption}
\usepackage{wrapfig}

\title{Steering Large Language Models using Conceptors: Improving Addition-Based Activation Engineering}

\author{%
  Joris Postmus\\
  University of Groningen\\
  Groningen, Netherlands \\
  \texttt{j.postmus@student.rug.nl} \\
  \And
  Steven Abreu\\
  University of Groningen\\
  Groningen, Netherlands \\
  \texttt{s.abreu@rug.nl}
}

\begin{document}

\maketitle
\vspace{-0.5cm}
\input{text/abstract.tex}
\vspace{-0.5cm}
\input{text/text-body.tex}

\clearpage

\bibliographystyle{unsrtnat}
\bibliography{ref}
% \bibliographystyle{unsrtnat}

%%%%%%%%%%%%%%%%%%%%%%%%%%%%%%%%%%%%%%%%%%%%%%%%%%%%%%%%%%%%

\clearpage
\appendix
\input{text/appendix.tex}

%%%%%%%%%%%%%%%%%%%%%%%%%%%%%%%%%%%%%%%%%%%%%%%%%%%%%%%%%%%%

% \include{checklisted}

\end{document}

%% file: text/abstract.tex
\begin{abstract}
Large language models have transformed AI, yet reliably controlling their outputs remains a challenge. 
This paper explores activation engineering, where outputs of pre-trained LLMs are controlled by manipulating their activations at inference time. Unlike traditional methods using a single steering vector, we introduce conceptors--mathematical constructs that represent sets of activation vectors as ellipsoidal regions. Conceptors act as soft projection matrices and offer more precise control over complex activation patterns. Our experiments demonstrate that conceptors outperform traditional methods across multiple steering tasks.
We further use Boolean operations on conceptors for combined steering goals that empirically outperform additively combining steering vectors on a set of tasks. These results highlight conceptors as a promising tool for more effective steering of LLMs. Our code is available on \href{https://github.com/jorispos/ConceptorSteering}{github.com/jorispos/conceptorsteering}.
\end{abstract}

%% file: text/text-body.tex
\section{Introduction}
\label{sec:introduction}
% \subsection{Motivation}
\label{sec:_motivation}

Large language models (LLMs) have rapidly advanced AI capabilities \cite{Xu2023Large}, but their potential to spread misinformation \cite{pan2023riskmisinformationpollutionlarge}, reinforce biases \cite{gallegos2024biasfairnesslargelanguage}, and develop harmful behaviors \cite{shevlane2023modelevaluationextremerisks} highlights the urgent need for methods to understand and control their outputs.
%In the context of this paper, 'steering' refers to the practice of reliably guiding a model's outputs toward or away from displaying the characteristics of a given pattern. A pattern can be specified through humanly interpretable examples, for example demonstrating concepts or behaviors (e.g. love, hate, etc.), but it can also include more complex types of outputs that may not as easily be directly described by humans.
%
Various methods, including reinforcement learning from human feedback (RLHF) \cite{Ouyang2022Training}, supervised fine-tuning \cite{devlin2019bertpretrainingdeepbidirectional}, and prompt engineering \cite{liu2021pretrainpromptpredictsystematic}, have been proposed to steer LLM outputs toward desired patterns. However, RLHF and fine-tuning are computationally expensive and struggle with generalization \cite{bottou2018optimizationmethodslargescalemachine, amodei2016concreteproblemsaisafety}, while prompt engineering often produces inconsistent results \cite{Chen2023Unleashing}.

\textit{Activation engineering} \cite{li2024inferencetimeinterventionelicitingtruthful, turner2024activationadditionsteeringlanguage} has recently been proposed as a new steering method which works by directly modifying the model's activations at inference time without changing the model's parameters and without expensive optimization.
A steering vector that represents desired behavior can be computed directly or (more commonly) contrastively from positive and negative examples \cite{panickssery2024steeringllama2contrastive}.
% It typically involves caching a set of token activation vectors from an LLM's forward pass on prompts that represent desired patterns (directly, e.g. ``up$\rightarrow$down'', or contrastively, e.g. ``love'' - ``hate''). These vectors are then subtracted or averaged to form a steering vector which can then be added onto a new forward pass to steer the model toward the desired pattern. This approach has shown to be effective at capturing and steering toward a wide range of patterns describing things like concepts (weddings, love, etc.) \cite{turner2024activationadditionsteeringlanguage}, functions (antonyms, synonyms, etc.) \cite{todd2024functionvectorslargelanguage}, and more complex behaviors (truthfulness, power-seeking, etc.) \cite{panickssery2024steeringllama2contrastive}.
However, finding contrastive prompts to identify complex patterns is not always possible and, more importantly, the performance of activation addition for steering is not reliable \cite{turner2024activationadditionsteeringlanguage}. 
% This may be explained by the inherent limitations that come with reducing the representation of a complex steering pattern into a single point (vector) in high-dimensional space. Compressing the representation of the pattern this much could lead to a loss of important steering information. 

This paper introduces an alternative to the predominant approach for steering LLMs using activation engineering. Instead of averaging or subtracting a set of activation vectors to form a steering vector, we use the cached activations to compute a \textit{conceptor} \cite{jaeger2014conceptorseasyintroduction}, which we refer to as a ``steering matrix''. Instead of manipulating the LLM's activations using vector addition, the activations are (softly) projected using a matrix-vector multiplication with the steering matrix.
%
% In this paper, we make the following contributions: (1) we introduce a novel conceptor-based steering mechanism for LLMs, (2) we use this method to improve function vectors (3) we extend the approach to Boolean combinations of function vectors, and (4) we use conceptor steering to reduce toxicity in LLMs.
We contribute the following: (1) we introduce a novel application of conceptors \cite{jaeger2014conceptorseasyintroduction} as steering mechanisms for LLMs, (2) we apply this mechanism to function vectors \cite{todd2024functionvectorslargelanguage} on GPT-NeoX and GPT-J, and (3) we show how a Boolean algebra on conceptors \cite{jaeger2017controllingrecurrentneuralnetworks} can be used for combining steering targets on GPT-J.

\section{Background}

Adding steering vectors to the residual stream has been used to control the output of LLMs across various domains \cite{turner2024activationadditionsteeringlanguage, panickssery2024steeringllama2contrastive, vanderweij2024extendingactivationsteeringbroad}. The use case that will mainly be focused on here are the findings from the paper by Todd \textit{et al.} \cite{todd2024functionvectorslargelanguage}. Their work showed that a steering vector can be extracted from the residual stream that captures the activation space of an input-output function (e.g. a function that takes a word and returns its antonym). This steering vector can then be added to the residual stream at inference time to steer the model toward performing the captured function. % For example, by prompting the model with “Hot” and adding the Antonym function vector to the residual stream during a new forward pass, the model would output “Cold”.
See Figure \ref{fig:fv_diagram} in Appendix \ref{app:exp-details-functions} for an illustration of function vector tasks.

\begin{figure}[t]
    \centering
    % \vspace{-1cm}
    \includegraphics[width=\textwidth]{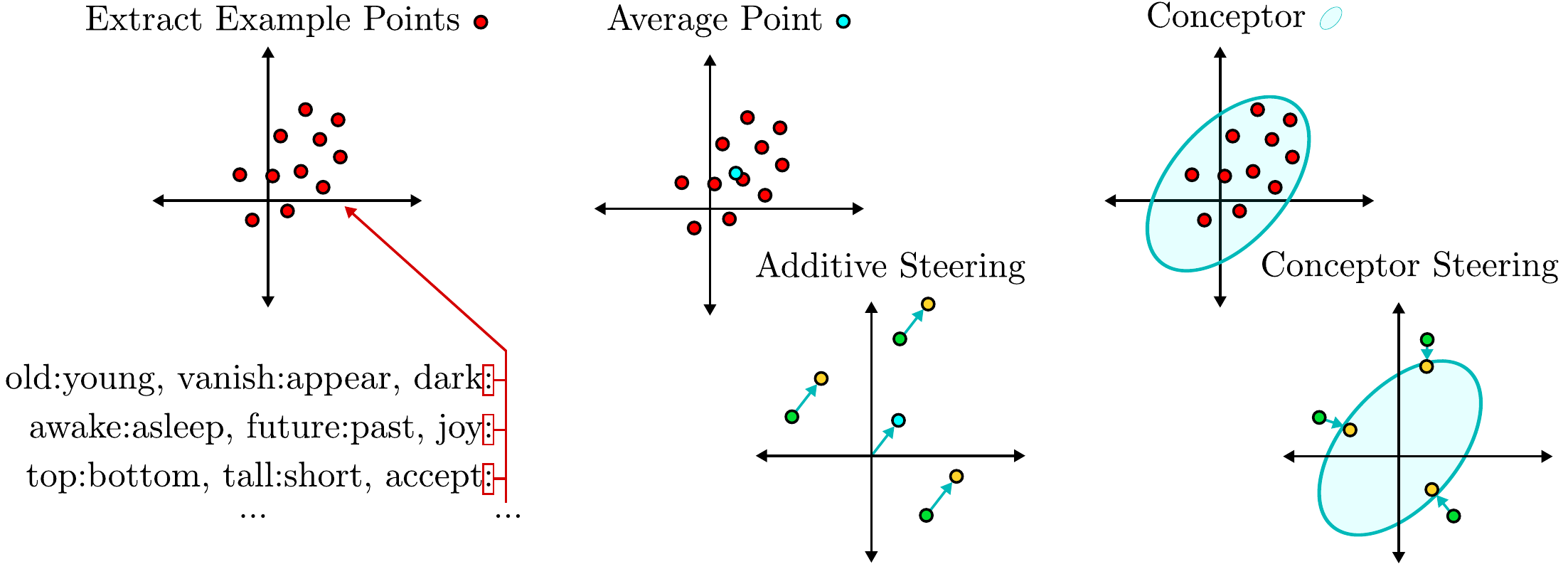}
    \caption{Illustration showing the basic geometric difference between additive and conceptor steering using a set of activations for the antonym task. Additive steering acts as a translation of the activation vectors by a fixed steering vector. Conceptor steering acts as a (soft) projection onto a target ellipsoid.}
    \label{fig:geometric-difference}
\end{figure}

Their baseline method works as follows. First, a set of in-context learning (ICL) prompts \( P_f \) that demonstrate a particular task \( f \) (the execution of an input-output function) are compiled. Then, for each prompt \( p_i^f \in P_f \) (\textit{e.g.,} $p_1^{antonym}=$ \texttt{hot:cold,old:}), the final token's activations \( h_{\ell}(p_i^f) \) are cached at a specific layer \( \ell \) from the residual stream \(h\). The cached activation vectors are then averaged into the steering vector \( \bar{h}_{\ell}^f \) for task \( f \) at layer \( \ell \):
\begin{equation}
\bar{h}_{\ell}^f = \frac{1}{|P_f|} \sum_{p_i^f \in P_f} h_{\ell}(p_i^f) 
\label{eq:function_vector_equation}
\end{equation}
To steer the model towards performing this function, the function (steering) vector \( \bar{h}_{\ell}^f \) can be added (without additional re-normalization) to the residual stream at layer \( \ell \) when the model would be completing a prompt containing a previously unseen input:
\begin{equation}
h_{\ell}' = \beta_{\text{add}} \, \bar{h}_{\ell}^{f} + h_{\ell}
\label{eq:additive-steering}
\end{equation}
where $h_{\ell}'$ is the steered activation and $\beta_{add}>0$ is a hyperparameter.
% This method has been used to steer the model toward executing a wide range of functions in a zero-shot context meaning that the model was not initially trained to do so, and was also not given explicit examples of this task in its prompt.
The performance of additive steering can further be improved by a technique called \textit{mean-centering} \cite{jorgensen2023improvingactivationsteeringlanguage}, see Appendix \ref{app:mean-centering}.

\section{Conceptors as Steering Matrices}

\begin{figure}[h!]
    \centering
    \includegraphics[width=0.8\textwidth]{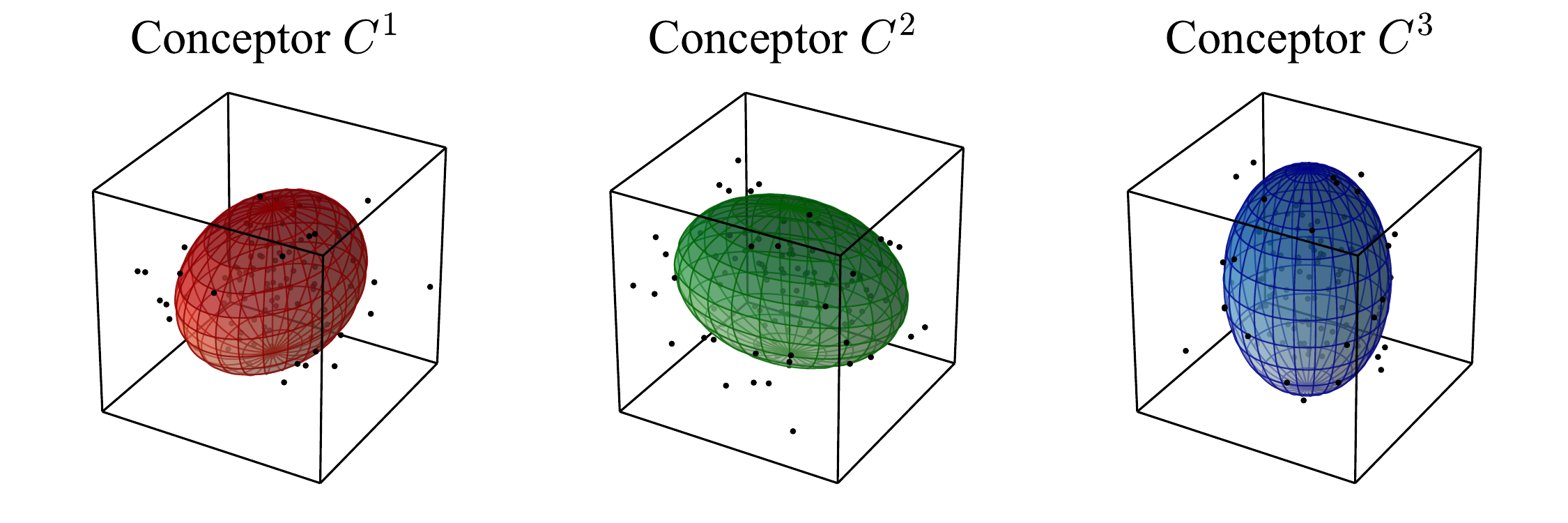}
    \caption{Illustration of three conceptors as ellipsoids that capture the state space region of different sets of neural activations in 3D space (black points). Reproduced from Jaeger \cite{jaeger2014conceptorseasyintroduction}.}
    \label{fig:conceptor-vis}
\end{figure}

Conceptors can broadly be defined as a neuro-computational mechanism designed to encapsulate and manipulate the state space of neural activations \cite{jaeger2014conceptorseasyintroduction}.
A conceptor matrix \(C\) is a positive semi-definite matrix that captures the principal directions and variances of a set of neural activation vectors. This structure can be visualized as a high-dimensional ellipsoid that describes the overall shape and spread of the activations' ``underlying pattern'', or state space region. See Figure \ref{fig:conceptor-vis} for a visual illustration.
Because conceptors are computed from the cloud of activation vectors and encode the correlations between activations, conceptors can better capture the activation space of complex patterns compared to simple point representations, which discard information about correlations. This difference is illustrated geometrically in Figure \ref{fig:geometric-difference}.

Conceptors have been used to control pattern-generating RNNs effectively across various behaviors \cite{jaeger2017controllingrecurrentneuralnetworks}, prevent catastrophic forgetting and enhance continual learning in feedforward networks \cite{2e7b9072e8704a16acda665979cf9e8a}, remove bias subspaces in LLMs like BERT and GPT \cite{yifei2023conceptoraideddebiasinglargelanguage}, and distill linguistic abstractions into knowledge graphs from contextual embeddings \cite{Kuiper2024, bricman2022}.

%%%% NEW ^^^ OLD VVV

One way to formalize the conceptor matrix \(C\), is through an optimization that minimizes the reconstruction error while using a regularization term. The objective function to be minimized is:
\[
\min_{C} \| X - XC \|^2_F + \alpha^{-2} \| C \|^2_F
\]
where \(X\) is a matrix of neural activation vectors (stacked as rows), \(\| \cdot \|_F\) is the Frobenius norm, and \(\alpha\) is the regularization parameter also referred to as the conceptor's \emph{aperture}. This aperture parameter \(\alpha\) balances the trade-off between accurately representing the activation pattern and maintaining a generalized representation.
The closed-form solution to this problem is given by:
\begin{equation}
C(R, \alpha) = R \left( R + \alpha^{-2} I \right)^{-1}
\quad \text{with} \quad
R = \frac{X^T X}{n}
\label{eq:conceptor_equation}
\end{equation}
where \(n\) is the number of samples, and \(I\) is the identity matrix of the same dimensionality as \(R\).

The eigenvalues \(\mu_i\) of the conceptor matrix \(C\) are defined as:
\[
\mu_i = \begin{cases}
    \frac{\lambda_i}{\lambda_i + \alpha^{-2}} & \text{for } 0 < \lambda_i < 1 \text{ and } 0 < \alpha < \infty \\
    0 & \text{for } 0 < \lambda_i < 1 \text{ and } \alpha=0 \\
    1 & \text{for } 0 < \lambda_i < 1 \text{ and } \alpha = \infty \\
    0 & \text{for } \lambda_i = 0 \text{ and } 0 \leq \alpha \leq \infty \\
    1 & \text{for } \lambda_i = 1 \text{ and } 0 \leq \alpha \leq \infty
\end{cases}
% \mu_i = \frac{\lambda_i}{\lambda_i + \alpha^{-2}},
\]
where \(\lambda_i\) represents the eigenvalues of the correlation matrix \(R\). These eigenvalues \(\mu_i\) fall within the interval \([0, 1]\) and are influenced by the aperture parameter \(\alpha\). 
When \(\alpha\) is large, the eigenvalues \(\mu_i\) approach 1 and \(C\) approaches the identity matrix, causing the conceptor to allow for more signal components to pass through the projection of the states with the conceptor matrix \(Cx\). Conversely, when \(\alpha\) is small, the eigenvalues \(\mu_i\) approach 0, causing the conceptor to allow for less variability. In the extreme case of \(\alpha \rightarrow 0\), the conceptor tends to the zero mapping.

We can use the conceptor for steering by collecting activations $h_\ell(p_i^f)$ into $X$ and then compute the associated conceptor $C_\ell^f$ using Equation \ref{eq:conceptor_equation}, and finally steer new hidden activations $h_\ell$ with:
\begin{equation}
h_{\ell}' = \beta_{c} C_\ell^f h_{\ell}
\label{eq:conceptor_steering}
\end{equation}
where $h_{\ell}'$ is the steered activation and $\beta_{c}>0$ is a hyperparameter. We can think of this as a ``soft projection''. A projection matrix has eigenvalues that are either zero or unity, but the conceptor matrix has ``soft'' eigenvalues between zero and unity. Thus, the conceptor ``softly projects'' the activation vector $h_{\ell}$ toward the pattern represented by $C_\ell^f$ by scaling its components according to the patterns' principal directions. 

\subsection{Boolean Operations on Conceptors}
\label{app:conceptors-boolean}

We can combine multiple steering matrices using the conceptor Boolean operations as defined by Jaeger \cite{jaeger2017controllingrecurrentneuralnetworks}. We begin with the OR operation on conceptors, which can be interpreted as merging the data from which each conceptor is computed by adding the covariance matrices on which $C_1$ and $C_2$ were computed. Given that $C_1$ was computed with the covariance matrix $R_1$ and $C_2$ was computed with the covariance matrix $R_2$, the conceptor that is computed on the sum of the two covariance matrices $R_1 + R_2$ is defined as $C_1 \vee C_2$:
\begin{align}
C_1 \vee C_2 &= (R_1 + R_2) (R_1 + R_2 + \alpha^{-2} I)^{-1} \\
C_1 \vee C_2 &= \left( I + \left(C_1 (I-C_1)^{-1} + C_2(I-C_2)^{-1}\right)^{-1} \right)^{-1}
\end{align}

The NOT operation on a conceptor $C$ is defined as the conceptor $\neg C$ that is computed on a covariance matrix $R^{-1}$ that is the inverse of the original covariance matrix $R$ for conceptor $C$. Intuitively, $\neg C$ can be interpreted as the conceptor that arises from data that which co-vary inversely to the data giving rise to $C$:
\begin{align}
\neg C &= R^{-1} (R^{-1} + \alpha^{-2} I)^{-1} \\
\neg C &= I - C
\end{align}

For our experiments, we use the AND operation which can now be obtained from the NOT and OR operations using de Morgan's law $a \wedge b = \neg (a \vee b)$ such that the conceptor $C_1 \wedge C_2$ is computed using the correlation matrix $(R_1^{-1} + R_2^{-1})^{-1}$. This leads to:
\begin{align}
C_1 \wedge C_2 &= (R_1^{-1} + R_2^{-1})^{-1} \left( (R_1^{-1} + R_2^{-1})^{-1} + \alpha^{-2} I \right)^{-1}\\
C_1 \wedge C_2 &= (C_1^{-1} + C_2^{-1} - I)^{-1}
\end{align}

% We can combine multiple steering matrices using the conceptor Boolean operations as defined by Jaeger \cite{jaeger2017controllingrecurrentneuralnetworks}By defining Boolean operations on conceptors, as done by Jaeger \cite{jaeger2017controllingrecurrentneuralnetworks}, we can combine multiple steering matrices using logical connectives. For thise present paper, we usemake use of the AND  operation which combiness two conceptors $C_1$ and $C_2$ into a joint conceptor $C_1 \wedge C_2$ or $C_1 \vee C_2$. See Appendix for \ref{app:conceptors-boolean} for more.with:
% \vspace{-0.5em}
% \begin{equation}
% C_1 \wedge C_2 = \left( C_1^{-1} + C_2^{-1} - I \right)^{-1}
% \label{eq:conceptor-and}
% \end{equation}
% The full derivation and rationale is given in Appendix \ref{app:conceptors-boolean}.

\subsection{Computational Complexity of Conceptor Steering}

The cost of computing a conceptor steering matrix is dominated by the matrix inversion and matrix-matrix multiplication of the activation correlation matrix $R=XX^T/n$ (see Equation \ref{eq:conceptor_equation}). This correlation matrix is a $n \times n$-dimensional matrix where $n$ is the dimension of the activation vectors (typically <4096 for the model sizes we presented, or up to 8192 for larger models such as Llama-2-70B), so the complexity of the conceptor computation is $\mathcal{O}(n^3)$. This computation is done entirely offline and the cost is amortized over all future applications of the steering method. 
The final conceptor $C \in \mathbb{R}^{n \times n}$ takes $O(n^2)$ memory -- the same amount as a weight matrix acting on the activation vectors. For 32-bit floating point numbers, this amounts to 17MB for $n=2048$, 67MB for $n=4096$, or 268MB for $n=8192$. 

During inference, conceptor steering adds an extra matrix-vector multiplication $Cx$ with the activation vector $x$.
However, the additional memory and inference cost for applying the conceptor can be eliminated by fusing the conceptor with the succeeding weight matrices for the query, key and value weight matrices. This is equivalent to replacing the existing weight matrix $W_x$ with the conceptor-fused weight matrix $W_x^C = W_x C$. This fusing of operations is standard practice when optimizing networks for inference. 
We note that there may be an overhead cost for switching the conceptor steering on and off which amounts to the cost of changing the network's computational graph during inference. We believe this overhead to be negligible during auto-regressive generation on a single data sample, but it must be considered when using batch sizes larger than one.

\section{Experiments}

For our experiments, we will use EleutherAI's GPT-J 6B and GPT-NeoX 20B models, as done in previous works on activation steering \cite{todd2024functionvectorslargelanguage,jorgensen2023improvingactivationsteeringlanguage}.
% For our experiments, we will make use of a decoder-only transformer neural network. More specifically, we will use \textit{EleutherAI’s GPT-J-6B} \cite{gpt-j} open-source 6 billion parameter model, pre-trained on the Pile dataset (825GB, $\sim$300B tokens) \cite{pile-dataset}. This was the model of choice by \cite{todd2024functionvectorslargelanguage} who showed it to be complex enough to have the representations of various input-output functions (of the kind shown in Section \ref{sec:__data}) encoded in its activation space.
For all experiments, we find optimal hyperparameters for each steering method at every layer. The details of our grid search for $\alpha$ and $\beta_c$ for conceptor-based steering and $\beta_{\text{add}}$ for additive steering can be found in Appendix \ref{app:hyperparameters}.

% The model contains a stack of 28 transformer layers, each with layer normalization, multi-head attention (MHA), and a feed-forward network (FFN). Central to the architecture is the residual stream\footnote{Throughout this paper, we will use the terminology from \cite{elhage2021mathematical}.}, which consists of a sequence of token activation vectors of shape \texttt{[num\_tokens, d\_model]}.
% Each token's embedded activation vector passes through the residual stream, where the MHA and FFN components sequentially add information to the residual stream at each transformer layer, with a normalization after each addition. The residual stream can thus be conceptualized as the network's "communication channel" \cite{elhage2021mathematical}.

\subsection{Function Steering}
\label{ss:function-steering}

We compare conceptor-based and additive steering mechanisms on their ability to steer a given model towards correctly executing a set of functions. We test both methods on GPT-J with 6B parameters and GPT-NeoX with 20B parameters. 
For each function, the described experiment will be repeated five times with different random seeds, and all reported results are averaged across across these five runs. 
The examples of the input-output functions come from the dataset by Todd \textit{et al.} \cite{todd2024functionvectorslargelanguage}. We use the following subset of five functions \cite{jorgensen2023improvingactivationsteeringlanguage}: antonyms (e.g. good$\rightarrow$bad), present-past (e.g. go$\rightarrow$went), English-French (e.g. hello$\rightarrow$bonjour), singular-plural (e.g. mouse$\rightarrow$mice), country-capital (e.g. Netherlands$\rightarrow$Amsterdam), and capitalize (e.g. word$\rightarrow$Word).
To ensure comparability of our results, we follow \cite{todd2024functionvectorslargelanguage} as closely as possible. For more details, see Appendix \ref{app:exp-details-functions}. 

\begin{figure}[h]
    \centering
    \includegraphics[width=1\textwidth]{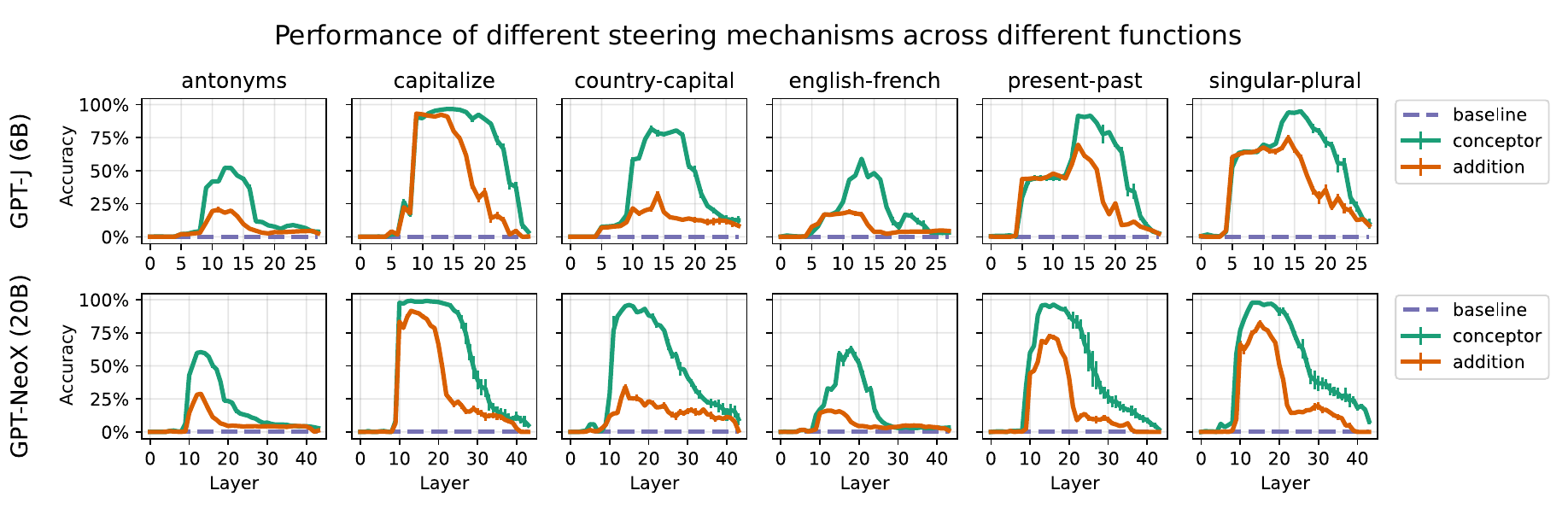}
    \caption{Comparison of the accuracy on all six function tasks for conceptor-based steering against additive steering across all layers for GPT-J and GPT-NeoX. For explanation, see main text.}
    \label{fig:fv_main_performance}
\end{figure}

The results in Figure \ref{fig:fv_main_performance} show that conceptor-based steering outperforms additive steering (the baseline method reported in Ref. \cite{todd2024functionvectorslargelanguage}) for every task on both tested models. 
% The performance improvement of the proposed conceptor mechanism is as large as 
In line with previous findings \cite{todd2024functionvectorslargelanguage,jorgensen2023improvingactivationsteeringlanguage}, steering is most effective across layers 9-16 for GPT-J and layers 10-30 for GPT-NeoX.

Table \ref{tbl:mc_gptj} and Figure \ref{fig:mc_gptj6} show that mean-centering (as outlined in Appendix \ref{app:mean-centering}) provides a small improvement for both addition-based and conceptor-based steering.
Mean-centering improves the performance of additive steering by as much as 2x (on the country-capital task). 
For conceptor-based steering the improvements of mean-centering are relatively smaller -- at most 5\% on the country-capital task. Conceptor-based steering outperforms additive steering on all tasks, even comparing additive steering with mean-centering against conceptor-based steering without mean-centering. 

\begin{table}[H]
\caption{The effect of mean centering on conceptor-based and addition-based steering on the GPT-J (6B) model, across simple function vector tasks. Results show the best performance across all hyperparameters and across all layers.}
\begin{tabular}{lccccc}
\toprule
 & antonyms & capitalize & country-capital & english-french & present-past \\
\midrule
Addition & 		 20.54\% & 93.16\% & 32.04\% & 18.88\% & 69.66\% \\
Addition (MC) &  31.20\% & 95.00\% & 63.90\% & 34.32\% & 83.32\% \\
Conceptor & 	 \underline{52.14\%} & \textbf{96.68\%} & \underline{81.62\%} & \underline{59.02\%} & \underline{91.56\%} \\
Conceptor (MC) & \textbf{52.82\%} & \underline{96.26\%} & \textbf{85.32\%} & \textbf{61.32\%} & \textbf{91.88\%} \\
\bottomrule
\end{tabular}
\label{tbl:mc_gptj}
\end{table}

\begin{figure}[H]
    \vspace{-0.4cm}
    \centering
    \includegraphics[width=\textwidth]{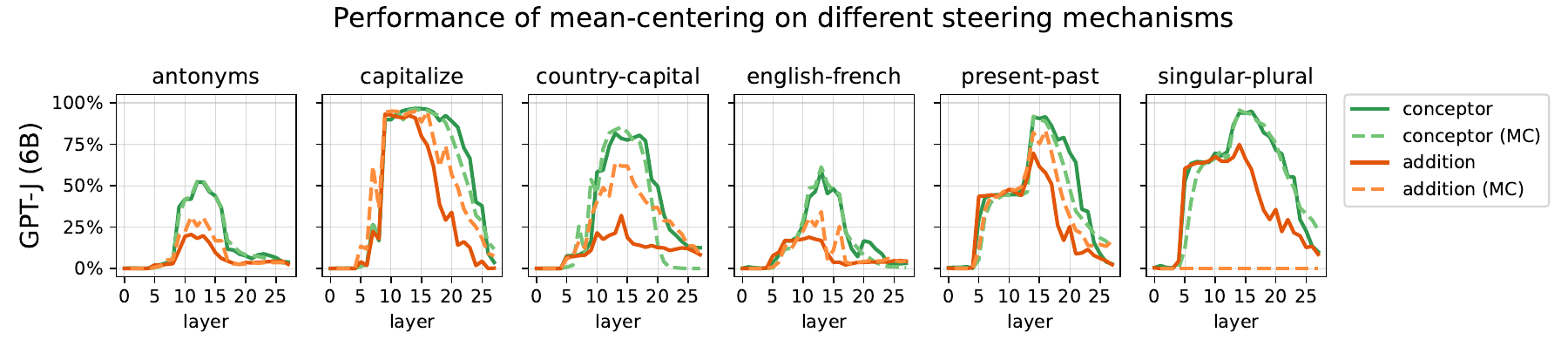}
    \caption{The effect of mean centering on conceptor-based and addition-based steering on the GPT-J (6B) model across all layers, computed on five different function vector tasks (\% accuracy). The line shows the best average performance across five runs for the best hyperparameters for the given layer.}
    \label{fig:mc_gptj6}
\end{figure}

\subsection{Steering Composite Functions}

We further conducted experiments where two conceptors, each representing one of three different functions, were combined using the AND operator. The input-output example dataset for this function was generated using GPT-4o. 
To present the baseline for how well non-combined steering mechanisms perform, we show results for the conceptor $C^{1,2}$ and the steering vector $\bar{h_\ell^{1,2}}$ that were each computed on the compound function directly. 
We then combine the conceptors computed on the individual functions $C^1$ and $C^2$ using the AND operation as $C^1 \wedge C^2$, and we combine the steering vectors $\bar{h_\ell^1}$ and $\bar{h_\ell^2}$ using their arithmetic mean $\frac{1}{2}(\bar{h_\ell^1} + \bar{h_\ell^2})$.

% \begin{wrapfigure}{lh!}{0.5\textwidth}
\begin{figure}[h]
  \centering
  \includegraphics[width=0.55\textwidth]{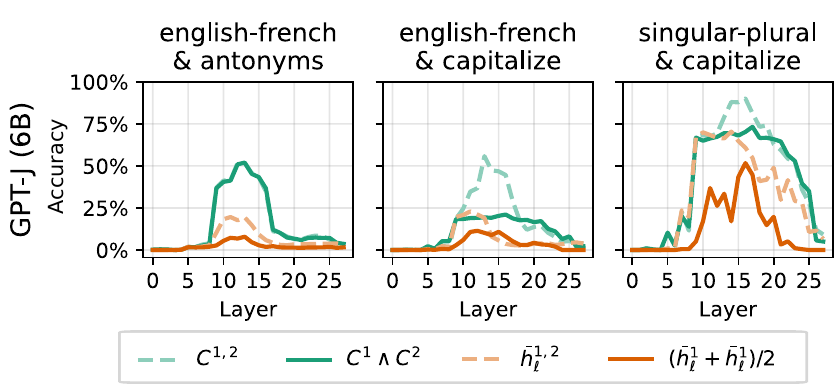}
  \caption{Performance of additive steering and conceptor steering on composite functions. For explanation of the figure caption, see text.
  Dashed lines represent the ``baseline'' where the steering mechanism is computed on the composite task. Solid lines show task arithmetic.
  % \vspace{-0.5cm}
  }
  \label{fig:fv_bool}
\end{figure}
% \end{wrapfigure}

Figure \ref{fig:fv_bool} shows the performance of all compared methods across all layers of the GPT-J model. 
In line with the results from Section \ref{ss:function-steering}, the conceptor baseline outperformed the additive baseline on all three tasks. The AND-combined conceptor outperformed the mean-combined steering vectors. On one of the three tasks, english-french \& antonyms, the AND-combined conceptor even outperforms the additive baseline.

\section{Conclusion}

In our experiments, conceptor-based steering generally outperformed addition-based methods. Further research should be conducted to assess the mechanisms' impact on the model's overall capabilities, the performance on more complex behaviors/tasks, and the scalability to larger models.

A limitation of conceptors is their reliance on more data points to build accurate representations. Additionally, the inherent mathematical structure and additional required computations makes it more computationally expensive compared to simple addition-based methods. However, while more expensive than addition-based approaches, they are still much cheaper than alternatives like RLHF and fine-tuning. 
Conceptors also introduce a new hyperparameter, the aperture $\alpha$, that may require tuning for optimal performance. In our experiments, we found a single aperture value, $\alpha = 0.1$, yields the best performance across all experiments\footnote{More precisely, the aperture value $\alpha=0.1$ is within $10\%$ of the best-performing aperture value across all experiments and models. In most experiments, it is the single best value. See Appendix \ref{app:hyperparam-sweep-results} for more details.}, but this finding must be verified for new models and steering tasks.

Despite these challenges, conceptor-based steering methods could offer a more precise and effective way to steer LLMs compared to traditional addition-based methods, proposing a fundamental shift in what is possible with activation engineering. 
Our experiments on conceptor-based steering further suggest that region-based representations may allow for more flexible and nuanced steering compared to point-based representations. The proposed method could have significant positive implications for debiasing models, aligning models with human values, and overall AI safety.

%% file: text/appendix.tex
\section{Appendix}

\subsection{Experimental Details}

All experiments were run on NVIDIA GPUs. The GPT-NeoX model was run on one NVIDIA RTX A6000 with 48GB of VRAM, and the GPT-J model was run on one NVIDIA GeForce RTX 4090 with 24GB of VRAM. Each hyperparameter sweep took less than 18 hours of compute time per model and per task. 

\subsubsection{Function Steering}
\label{app:exp-details-functions}

All the experimental configurations (number of experiments, number of ICL prompts and examples per prompt, accuracy metric, etc.) were, unless mentioned otherwise, adopted from Ref. \cite{todd2024functionvectorslargelanguage} to ensure comparability of results.

\begin{figure}[h!]
    \centering
    \begin{subfigure}[t]{0.55\columnwidth}
        \centering
        \includegraphics[width=0.55\linewidth, keepaspectratio=true]{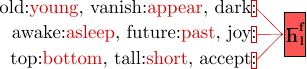}
        \caption{Extraction of the antonym function (steering) vector \( \bar{h}_{\ell}^f \) at layer \(l\) using 3 ICL prompts.}
        \label{fig:left_diagram}
    \end{subfigure}
    \hfill
    \begin{subfigure}[t]{0.38\columnwidth}
        \centering
        \includegraphics[width=0.55\linewidth, keepaspectratio=true]{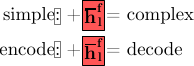}
        \caption{Antonym steering vector in 2 zero-shot contexts.}
        \label{fig:right_diagram}
    \end{subfigure}
    \caption{Visualization of how an antonym function (steering) vector can be extracted and applied. Example from \cite{todd2024functionvectorslargelanguage}}.
    \label{fig:fv_diagram}
\end{figure}

For each experiment, to generate the 4 steering mechanisms, we first compile \( N_p = 100 \) (ICL) prompts that demonstrate the respective input-output function. The prompts are formed by randomly sampling \( N = 10 \) input-output pairs from the function pairs dataset. If for a specific function, the dataset contains less than \( N_p \times N = 1000 \) input-output examples, this sampling is done with replacement. For each prompt \( p_{i}^{f} \), the last input-output pair has the output stripped, resulting in the format:

\[ p_i^f = "x_1:y_1, x_2:y_2, ..., x_{N-1}:y_{N-1}, x_N:" \]

where \( x \) represents the input tokens of a randomly sampled (input, output) pair, \( y \) represents the corresponding output tokens, \(N\) represents the number of sampled input-output pairs, and \(i \in \{1, \ldots, N_p\}\). A very simple example where \(N_p=3\) and \(N=3\) can be seen in Figure \ref{fig:left_diagram}. 

Formally, for each function \( f \in F \) in our set of in-context learning (ICL) tasks, we have compiled a set \( P_f \) of ICL prompts \( p_{i}^{f} \in P_f \). Each prompt \( p_{i}^{f} \) is a sequence of tokens with \( N \) input-output exemplar pairs \((x, y)\) that demonstrate the function \( f \) mapping between \( x \) and \( y \). For each experiment, we generate \( N_p \) such prompts.

Now that the ICL prompts have been generated, we need to extract the relevant activations. Todd \textit{et al.} \cite{todd2024functionvectorslargelanguage} showed that the neural representations of the functions are encoded in the activation vector of the last token (":") of the prompt, right before the transformer would auto-regressively start generating the output token(s). Moreover, the point in the residual stream \(h\) at which the functions were most strongly encoded was shown to be at the beginning of layers \( L = \{9, \ldots,16\} \), right before MHA and FFN \cite{todd2024functionvectorslargelanguage}.

Formally, for each function \( f \in F \) and each prompt \( p_{i}^{f} \in P_f\), the activation vectors \( h_{\ell}^{f}(p_{i}^{f}) \) are extracted from the residual stream \(h\) at each relevant layer \( l \in L \) from the last token's (":") activation vector.

For each function \( f \in F \) and each layer \( l \in L \), we now have \( N_p \) cached activation vectors \( h_{\ell}^{f}(p_{i}^{f}) \) aimed to encode the neural representation of \(f\) at layer \(l\). Using this, we can generate the layer-specific steering mechanisms for each function as follows:
\begin{itemize}
  \item The standard additive steering mechanism \(\bar{h}_{\ell}^{f}\) is generated by averaging over all the cached activation vectors \( h_{\ell}^{f}(p_{i}^{f}) \) respectively as described in Equation \ref{eq:function_vector_equation}.

 \item The additive steering mechanism with mean-centering \(\bar{h}_{\ell}^{f,\text{mc}}\) is computed by taking the previously generated steering mechanism \(\bar{h}_{\ell}^{f}\) and subtracting \(\mu_{\text{train}}\) as described in Equation \ref{eq:mean_centering_equation}.

\item The regular conceptor steering mechanism \( C \) is computed as described in Equation \ref{eq:conceptor_equation} using the aperture value \(\alpha_{\text{reg}}\). The correlation matrix \( R \) is computed as \( R = \frac{X^T X}{N_p} \), where \( X \) is the matrix of all \( h_{\ell}^{f}(p_{i}^{f}) \) stacked activation vectors.

\item The mean-centered conceptor steering mechanism \(C^{\text{mc}} \) is computed with some minor adjustments. The matrix \( X \) is formed by subtracting \(\mu_{\text{train}}\) from the activation vectors \( h_{\ell}^{f}(p_{i}^{f}) \) before stacking them. This results in an adjusted correlation matrix \(R\):

\[
R = \frac{(X - \mu_{\text{train}})^T (X - \mu_{\text{train}})}{N_p}
\]

The mean-centered conceptor matrix \(C^{\text{mc}} \) can then be calculated as described in Equation \ref{eq:conceptor_equation} using the aperture value \(\alpha_{\text{mc}}\) and the adjusted correlation matrix \(R\).
\end{itemize}

To test the performance of the generated steering mechanisms, new sets of \( N_t = 1000 \) input-output pairs are randomly sampled from the function pairs dataset for each experiment. This is done with replacement for functions where the dataset contains less than \(N_t\) pairs. An input prompt \(p_t\) is formatted as \(p_t = "x:"\), where \(x\) is a tokenized input from an input-output pair. The tokenized output \(y\) from the pair is left out from \(p_t\) as it will be used to test the accuracy of the steering mechanisms. For each experiment, we now have \(N_t\) test input prompts \(p_t\).

To test the accuracy of the steering mechanisms, we apply the layer-specific steering mechanisms on independent forward passes and record their subsequent output. This means that for our experimental configuration, across the functions \(f \in F\), the 5 experiments, the 4 steering mechanisms (excluding the baseline), the \(N_t\) number of test prompts, and the number of layers \(l \in L\), there will be \(6 \times 5 \times 4 \times 1000 \times 8 = 960,000\) forward passes, each with a steering intervention. 

Each steering intervention will consist of a layer-specific steering mechanism modifying the residual stream \(h\) at the mechanisms' respective layer \(l\). This modification can be defined as transforming the unmodified residual stream activation vector \(h_{\ell}\) into the steered activation vector \(h_{\ell}'\). The steering mechanisms' modification can be described as follows:

\begin{itemize}
\item For the standard additive steering mechanism, the averaged activation vector \(\bar{h}_{\ell}^{f}\) is multiplied by the injection coefficient \(\beta_{\text{add}}\) and added to the residual stream activation vector \(h_{\ell}\):
\[
h_{\ell}' = \beta_{\text{add}} \, \bar{h}_{\ell}^{f} + h_{\ell}
\]
\item For the additive steering mechanism with mean-centering, the mean-centered average activation vector \(\bar{h}_{\ell}^{f,\text{mc}} \) is multiplied by the injection coefficient \(\beta_{\text{add}}\) and added to the residual stream activation vector \(h_{\ell}\):
\[
h_{\ell}' = \beta_{\text{add}} \, \bar{h}_{\ell}^{f,\text{mc}} + h_{\ell}
\]
\item For the regular conceptor steering mechanism, the residual stream activation vector \(h_{\ell}\) is multiplied using the conceptor matrix \( C \) and further multiplied with the rescaling coefficient \(\beta_{\text{c}}\):
\[
h_{\ell}' = \beta_{\text{c}} \, C \, h_{\ell}
\]
\item For the mean-centered conceptor steering mechanism, the residual stream activation vector \(h_{\ell}\) is first adjusted by subtracting \(\mu_{\text{train}}\). This adjusted vector is then multiplied with the mean-centered conceptor matrix \(C^{\text{mc}}\) and further multiplied with the rescaling coefficient \(\beta_{\text{c}}\). Finally, \(\mu_{\text{train}}\) is added back to the result:
\[
h_{\ell}' = \beta_{\text{c}} \, C^{\text{mc}} \, (h_{\ell} - \mu_{\text{train}}) + \mu_{\text{train}}
\]
\item For the baseline condition, no modifications are made to the residual stream.
\[
h_{\ell}' = h_{\ell}
\]

\end{itemize}
After the respective modifications have been made to the residual stream, the forward passes will continue as usual. At the end of each forward pass, the final logits are converted into probabilities using a softmax, and the token with the highest probability is selected. This means that at the end of one experiment, we have \(N_t\) single-token outputs for each layer-specific steering mechanism. These tokens can now be compared with the first token of output \(y\) that corresponds with the input \(x\) of the initial prompt \(p_t\). Based on how many of the \(N_t\) outputs were correctly identified, a top-1 accuracy is calculated for each layer-specific steering mechanism. This experiment is repeated 5 times for each function \( f \in F \) to account for variability caused by the random sampling for the generation of the steering mechanisms and test sets. 

\subsubsection{Hyperparameter optimization}
\label{app:hyperparameters}

The performance of the steering mechanisms in the function vector experiments was optimized through a grid search over all hyperparameters. Firstly, we try steering at each layer of the model. For conceptor-based steering, we do a grid search for the aperture value $\alpha$ with possible values from $\{0.001, 0.0125, 0.05, 0.1\}$ and the scaling coefficient $\beta_c$ with possible values from $\{0.5, 1.0, 2.0, 3.0, 4.0, 5.0\}$. For additive steering, we run a grid search over the scaling coefficient $\beta_{\text{add}}$ with possible values from $\{0.5, 1.0, 1.5, 2.0, 2.5, 3.0, 4.0, 5.0\}$. The results from these hyperparameter sweeps are shown in Appendix \ref{app:hyperparam-sweep-results}

\subsection{Additional Experimental Results}

\subsubsection{Mean centering}
\label{app:mean-centering}

An important improvement for additive steering is a technique called \textit{mean-centering}, put forward by Jorgensen \textit{et al.} \cite{jorgensen2023improvingactivationsteeringlanguage}. This method enhances the effectiveness of steering vectors by reducing the inherent bias present in the activation space of LLMs. Activation vectors in LLMs tend to be anisotropic, meaning that they are not evenly distributed around the origin, but are instead offset in a consistent direction. This can negatively impact the steering vector's performance as the bias vector \( b \) representing this offset, does not encode any specific task-related information, diluting the steering vector’s effectiveness.

First, the steering vector \( \bar{h}_{\ell}^f \) for a specific function \( f \) is computed by averaging the activations at layer \( \ell \) on a set of ICL prompts demonstrating the input-output function \( P_f \) (as defined in Equation \ref{eq:function_vector_equation}).

\( \bar{h}_{\ell}^f \) now encodes the task-specific behavior but may still be affected by biases in the model’s overall activation space. Mean-centering attempts to mitigate this by subtracting the mean activation of a broader dataset that represents the general activation space of the model. This is done by computing the mean activation vector \( \mu_{\text{train}} \) over a large, representative set of prompts \( D_{\text{train}} \) from the model’s training data.

% TODO: @joris, check that this is correct for our setup since we're doing the mean centering experiments only on the GPT-J model now. 
The mean activation vector \( \mu_{\text{train}} \) was calculated using the same procedure described by Jorgensen \textit{et al.} \cite{jorgensen2023improvingactivationsteeringlanguage}: A subset from the dataset used to train GPT-2 was compiled \cite{Gokaslan2019OpenWeb}.
The subset was constructed by storing all entries from the folders \texttt{urlsf\_subset01-1/data} and \texttt{urlsf\_subset01-182/data}.
After this, only entries that contained less than 500 tokens (using the GPT-2 Tokenizer) were retained. This resulted in 210 entries from which the final 10 were removed, leaving a dataset of 200 entries. The mean activation vector \( \mu_{\text{train}} \) was then computed by averaging the activations over this dataset.

Implementing the mean-centering performance enhancement for steering toward the execution of functions can be done as follows:

\begin{equation}
\bar{h}_{\ell}^{f,\text{mc}} = \bar{h}_{\ell}^f - \mu_{\text{train}} 
\quad \text{with} \quad
\mu_{\text{train}} = \frac{1}{|D_{\text{train}}|} \sum_{d \in D_{\text{train}}} h_{\ell}(d) 
\label{eq:mean_centering_equation}
\end{equation}

where $\bar{h}_{\ell}^f$ is as described in Equation \ref{eq:function_vector_equation}, and $D_{\text{train}}$ is the dataset for which the mean-centered vector $\mu_{\text{train}}$ is computed.
This refinement leads to a steering vector that can more effectively guide the model toward the specific task and has been shown to have a positive impact on the overall steering effectiveness \cite{jorgensen2023improvingactivationsteeringlanguage}.

\subsection{Hyperparameter Sweep Results}
\label{app:hyperparam-sweep-results}

In the following section, we present results from the hyperparameter optimization described in Appendix \ref{app:hyperparameters}, in order to assess the sensitivity of both steering mechanisms (additive and conceptor-based) to the hyperparameters. 

\subsubsection{Conceptor Steering}

\begin{figure}[h]
    \centering
    \includegraphics[width=\textwidth]{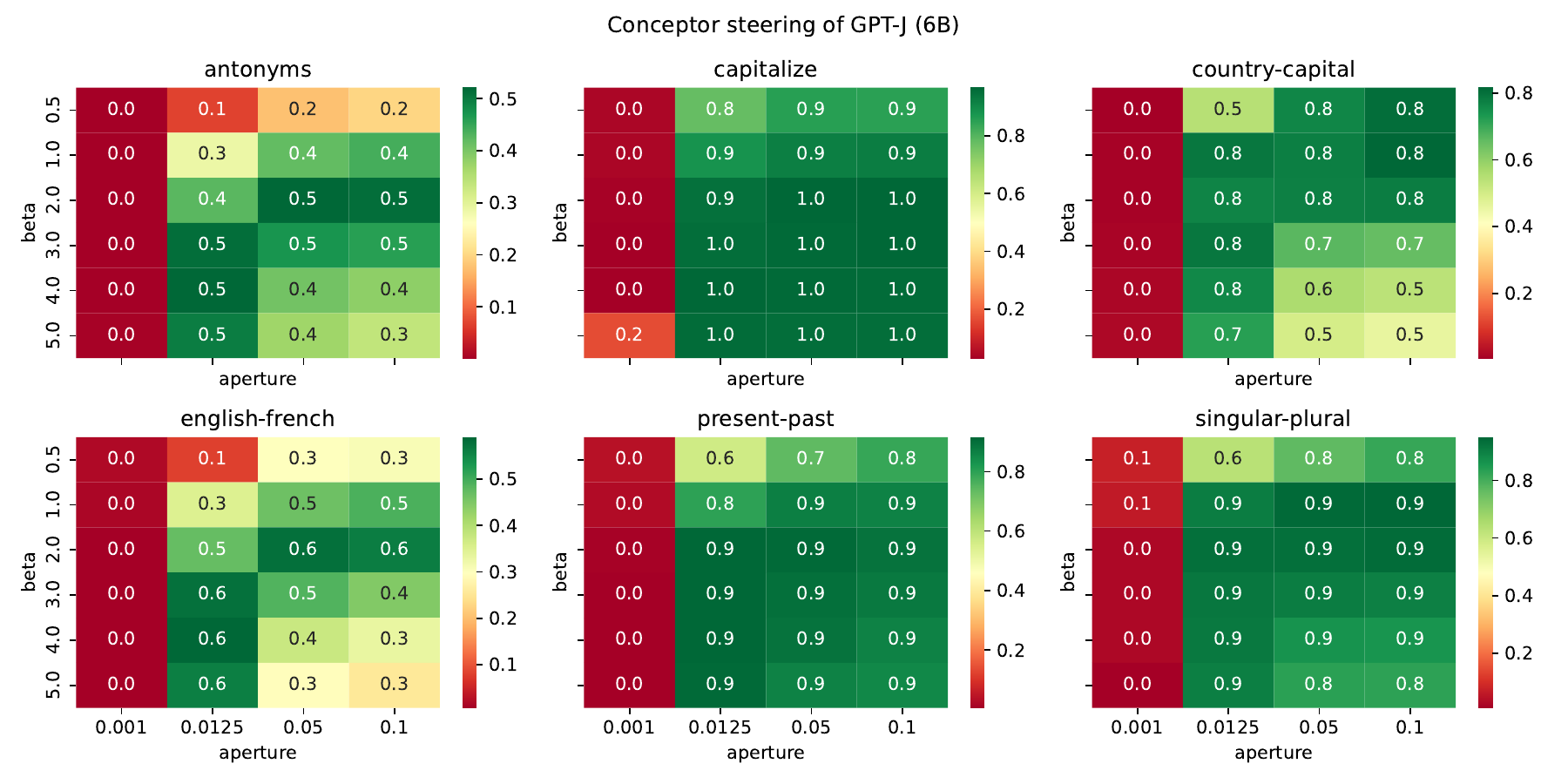}
    \caption{Performance results of the grid search across aperture and beta values (for the optimal layer) for the GPT-J (6B) model, using conceptor-based steering.}
    \label{fig:hp-conceptor-gptj-layer}
\end{figure}

\begin{figure}[h]
    \centering
    \includegraphics[width=\textwidth]{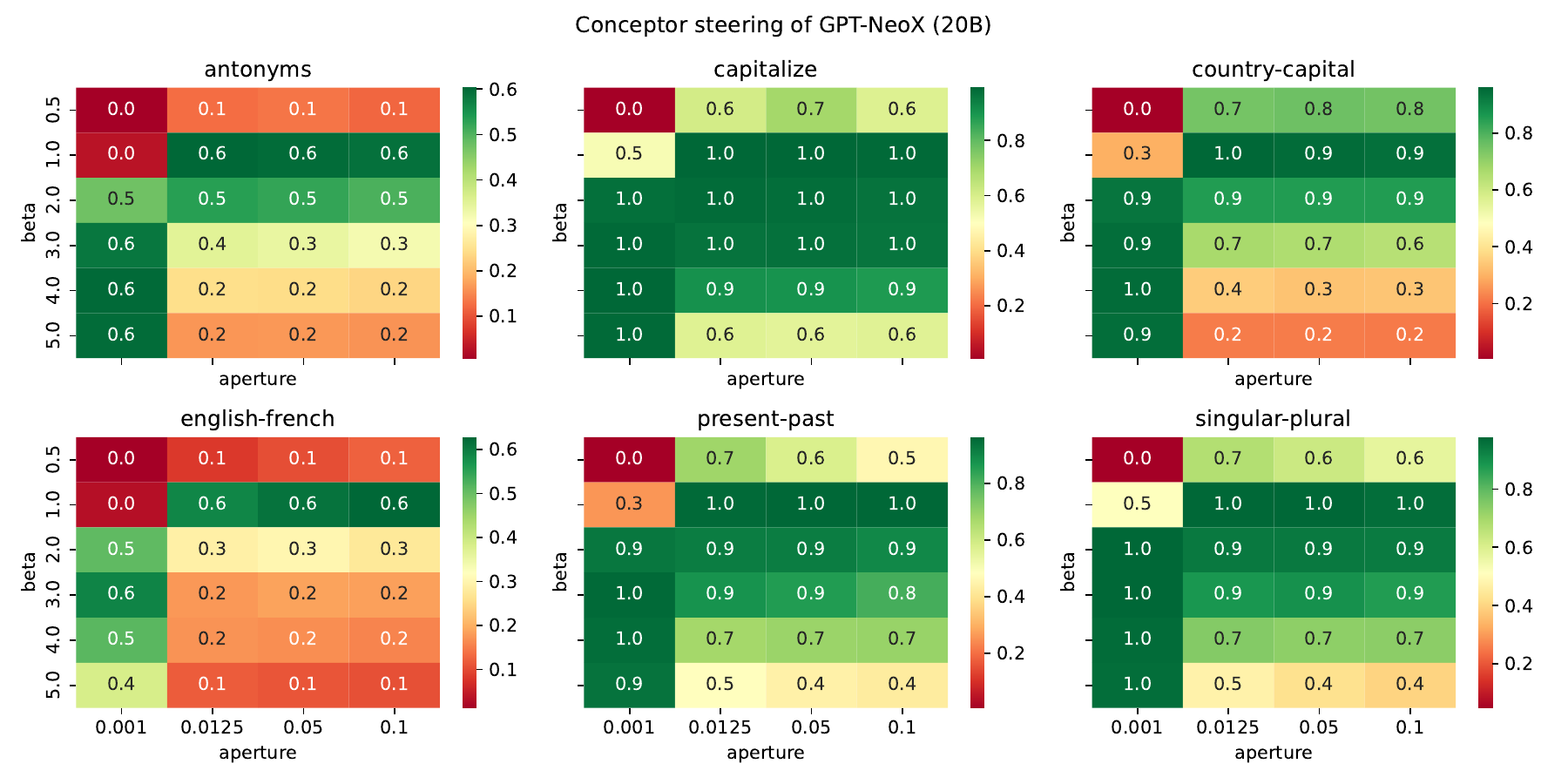}
    \caption{Performance results of the grid search across aperture and beta values (for the optimal layer) for the GPT-NeoX (20B) model, using conceptor-based steering.}
    \label{fig:hp-conceptor-neox-layer}
\end{figure}

Figure \ref{fig:hp-conceptor-gptj-layer} shows that the optimal choice of aperture and beta parameters for the conceptor steering mechanism is constant at $\alpha=0.05$ and $\beta_C=2.0$ across all tasks for the GPT-J model (for the layer with the maximum performance). Figure \ref{fig:hp-conceptor-neox-layer} shows similar behavior for the GPT-NeoX model, although the optimal beta parameter is $\beta=1$ and the optimal aperture parameter changes to $\alpha=0.0125$ for the country-capital task, and $\alpha=0.1$ for the english-french task, and $\alpha=0.05$ for all other tasks. 
This shows that hyperparameter choices are robust for conceptor steering, but still benefit from task-specific and model-specific optimization. 

\begin{figure}[h]
    \centering
    \includegraphics[width=\textwidth]{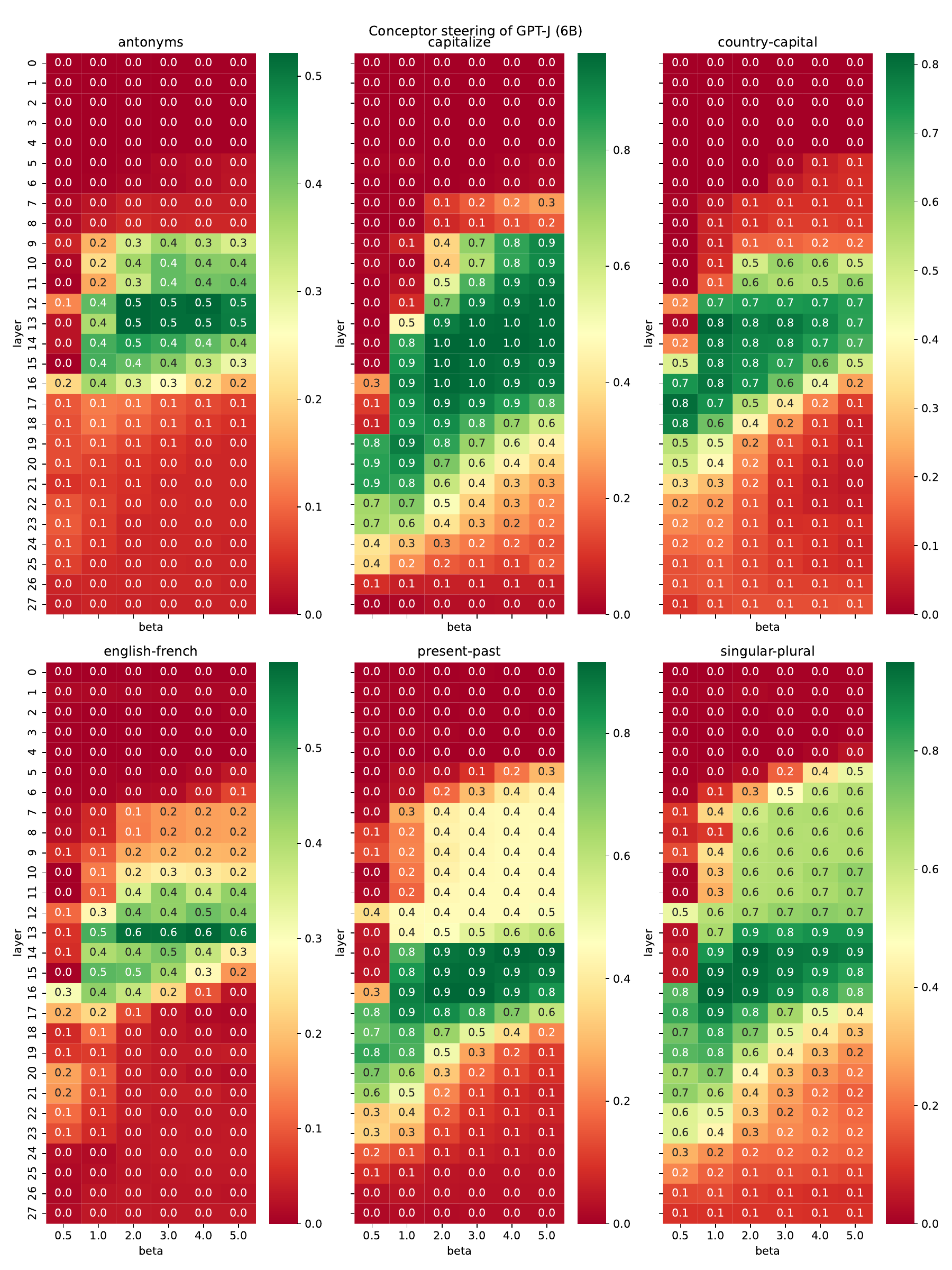}
    \caption{Performance results of the grid search across layers and beta values (for the optimal aperture value) for the GPT-J (6B) model, using conceptor-based steering.}
    \label{fig:hp-conceptor-gptj-aperture}
\end{figure}

\begin{figure}[h]
    \centering
    \includegraphics[width=\textwidth]{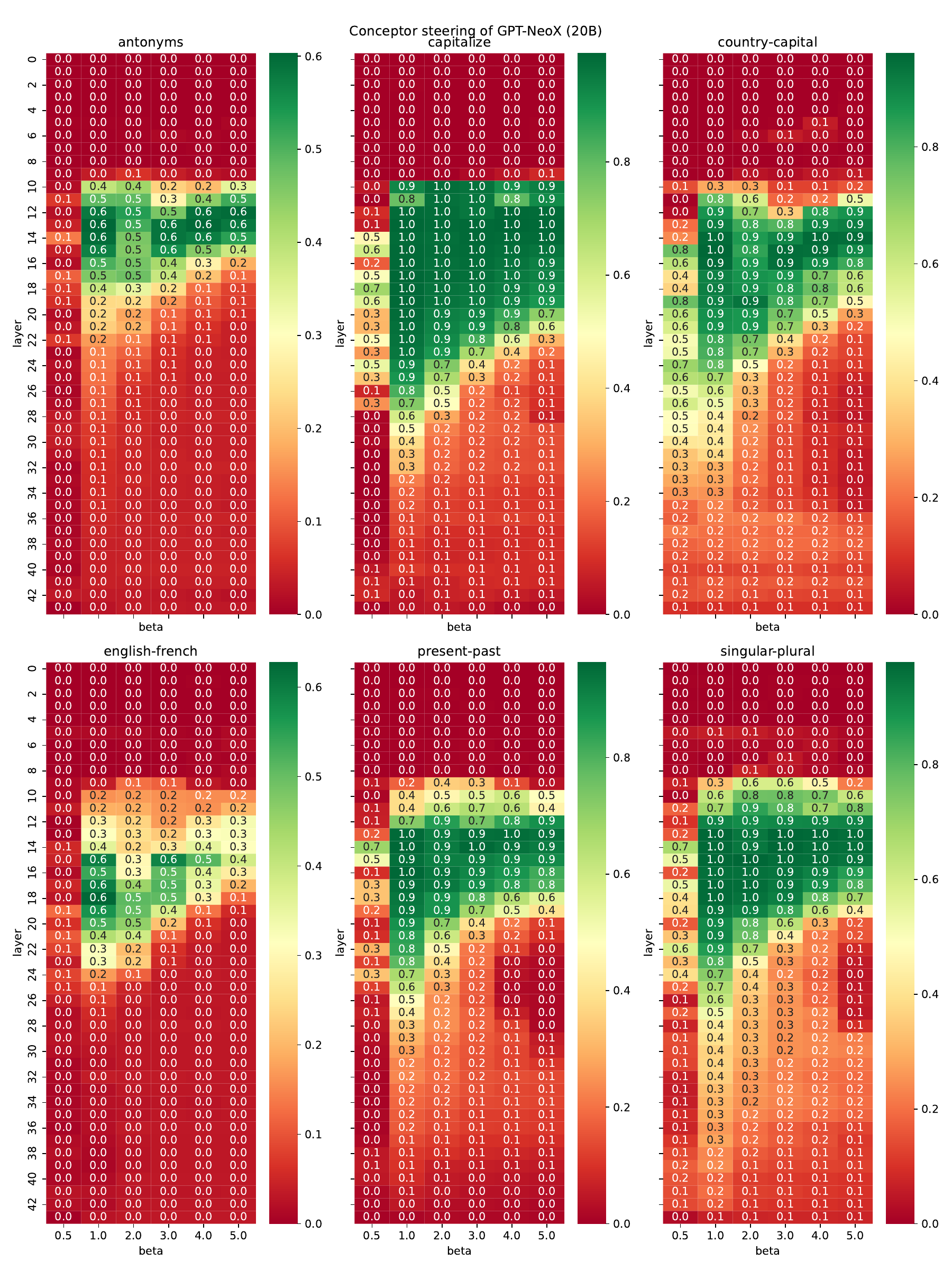}
    \caption{Performance results of the grid search across layers and beta values (for the optimal aperture value) for the GPT-NeoX (20B) model, using conceptor-based steering.}
    \label{fig:hp-conceptor-neox-aperture}
\end{figure}

\begin{figure}[h]
    \centering
    \includegraphics[width=\textwidth]{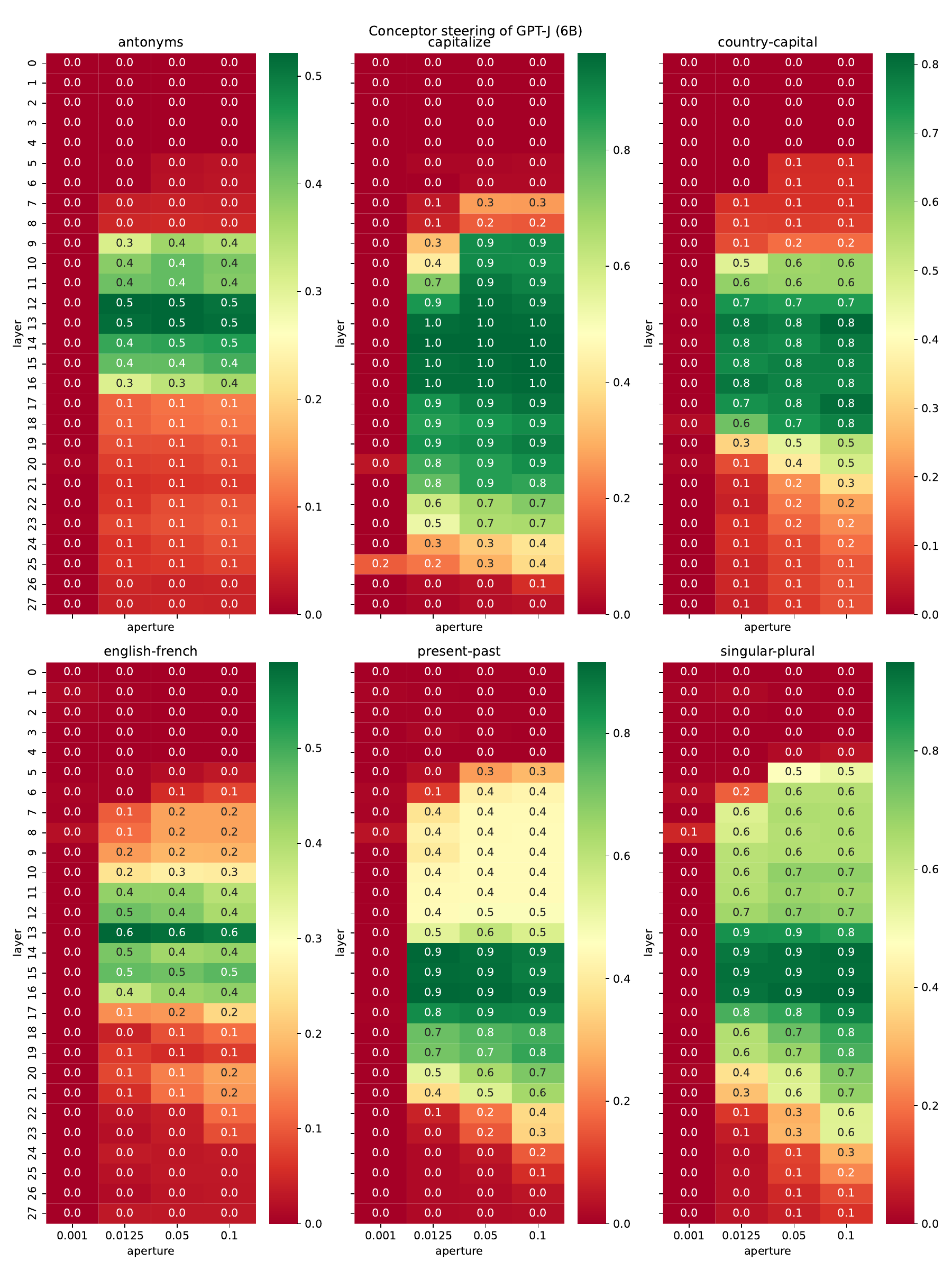}
    \caption{Performance results of the grid search across layers and aperture values (for the optimal beta value) for the GPT-J (6B) model, using conceptor-based steering.}
    \label{fig:hp-conceptor-gptj-beta}
\end{figure}

\begin{figure}[h]
    \centering
    \includegraphics[width=\textwidth]{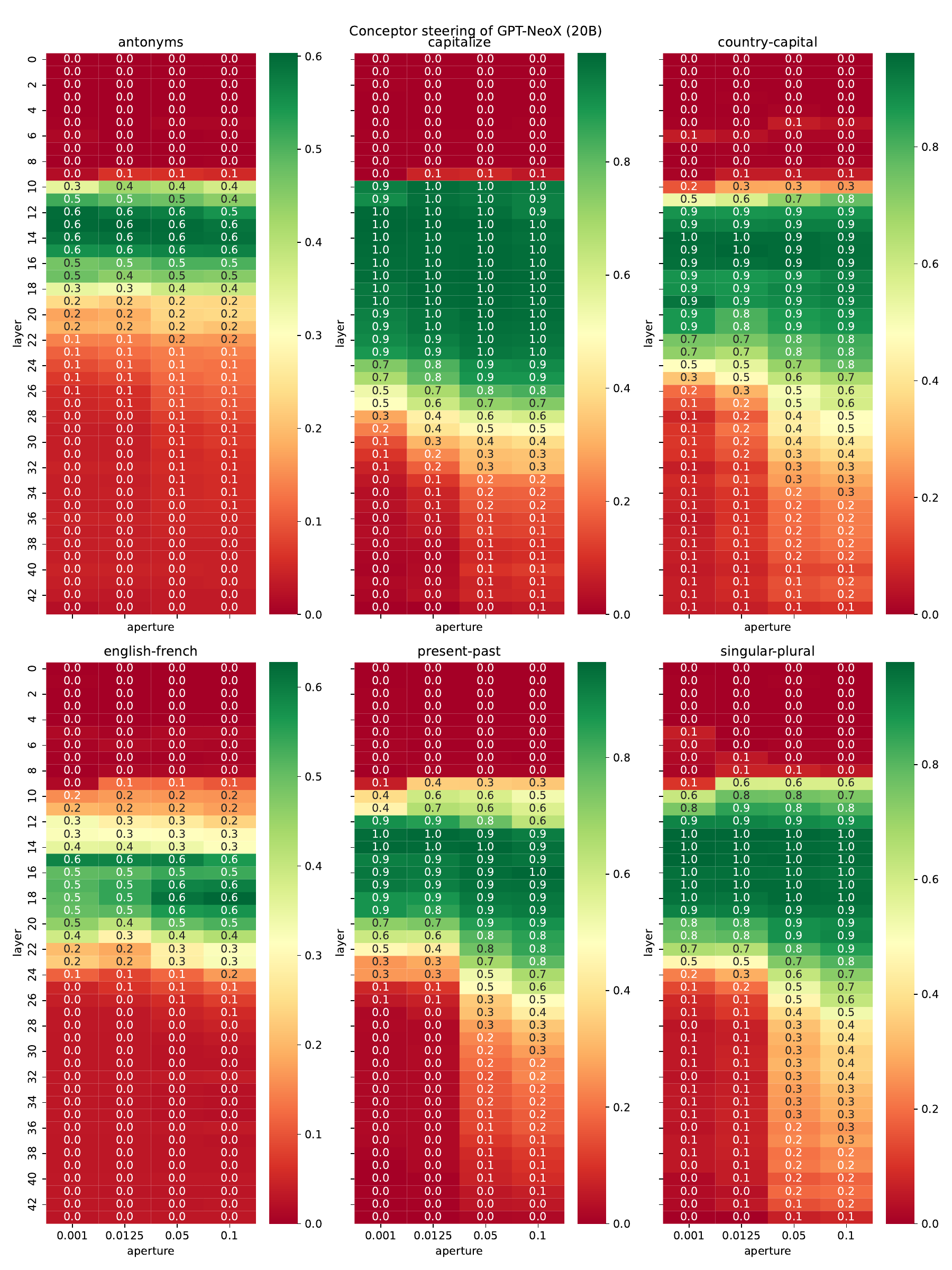}
    \caption{Performance results of the grid search across layers and aperture values (for the optimal beta value) for the GPT-NeoX (20B) model, using conceptor-based steering.}
    \label{fig:hp-conceptor-neox-beta}
\end{figure}

We further show the performance of conceptor-based steering across all layers and different beta values (taking the best-performing aperture value) for the GPT-J model in Figure \ref{fig:hp-conceptor-gptj-aperture} and for the GPT-NeoX model in Figure \ref{fig:hp-conceptor-neox-aperture}. For the GPT-J model, the best-performing layers are typically layers 12-14 with some variability (present-past being a few layers later at 14-17, and capitalize working well across layers 9-19). For the GPT-NeoX model, conceptor steering reaches (near-)maximum performance at layer 15 across all tasks, with layer 15 being at around one third of the depth of the model.
Figures \ref{fig:hp-conceptor-gptj-beta} and \ref{fig:hp-conceptor-neox-beta} show the performance of conceptor-based steering across all layers and different aperture values (taking the best-performing beta value) for the GPT-J model and the GPT-NeoX model, respectively, and show a similar pattern as described above.

\subsubsection{Additive Steering}

\begin{figure}[h]
    \centering
    \includegraphics[width=\textwidth]{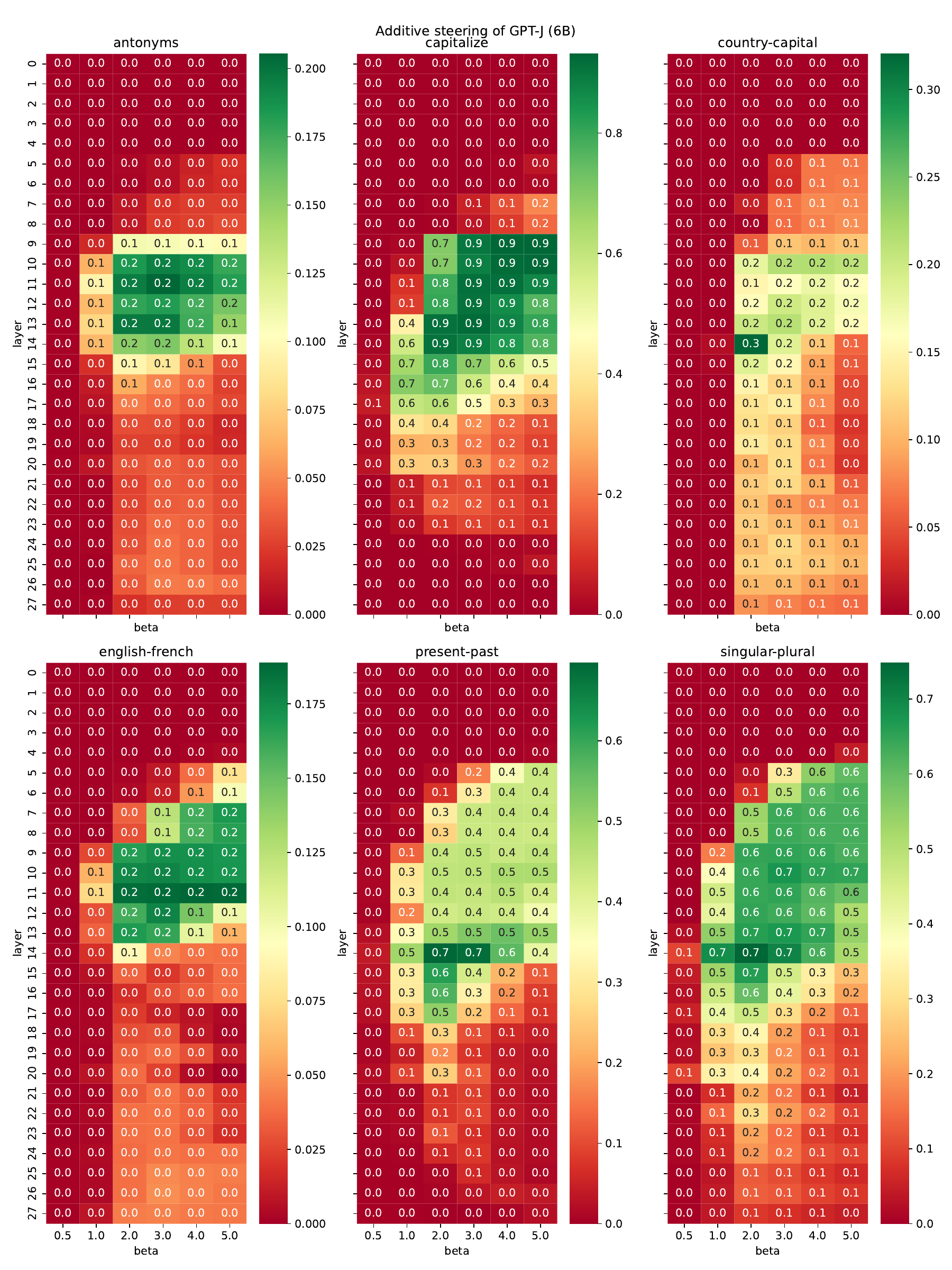}
    \caption{Performance results of the grid search across layers and beta values for the GPT-J (6B) model, using additive steering.}
    \label{fig:hp-addition-gptj}
\end{figure}

\begin{figure}[h]
    \centering
    \includegraphics[width=\textwidth]{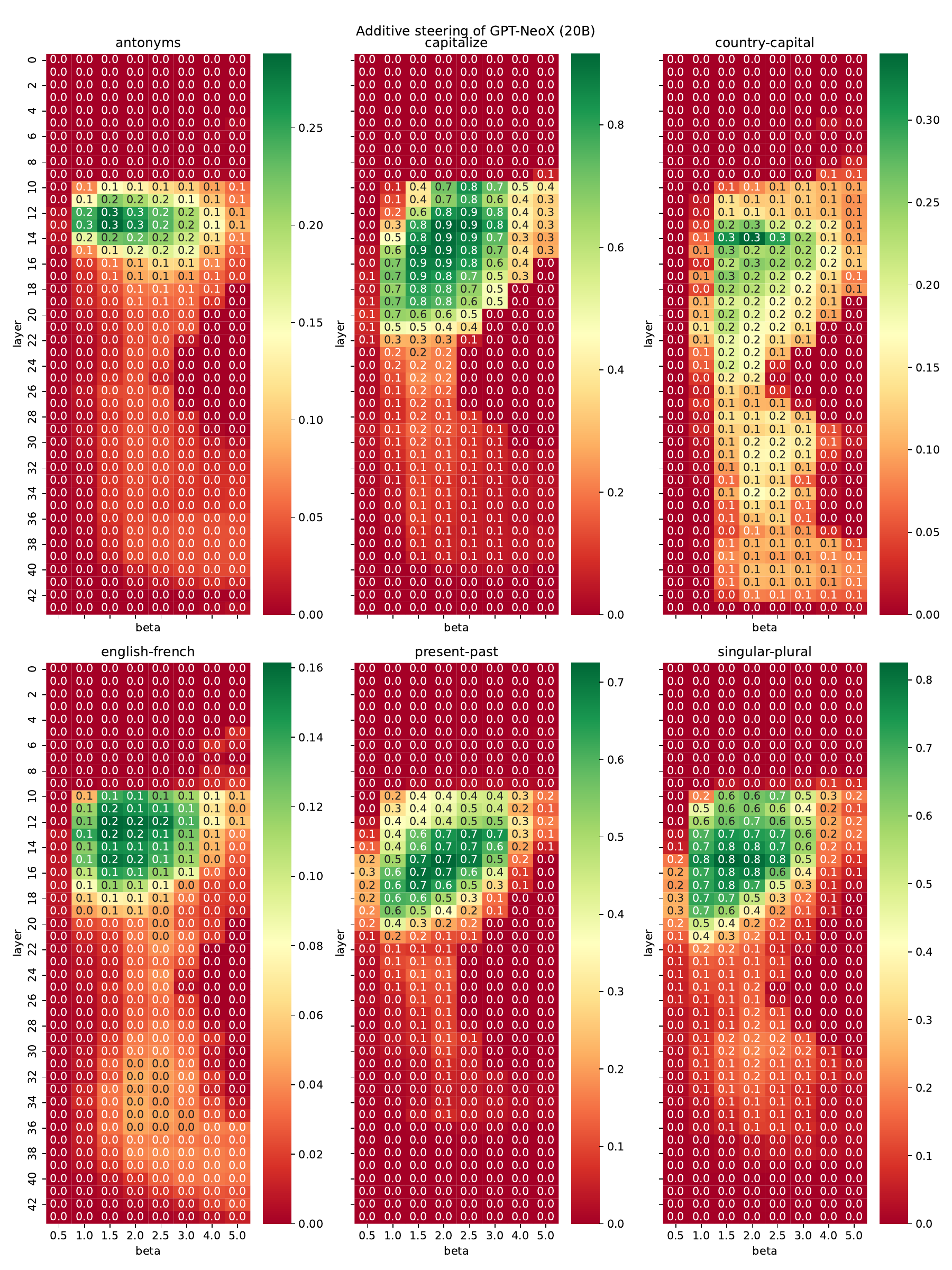}
    \caption{Performance results of the grid search across layers and beta values for the GPT-NeoX (20B) model, using additive steering.}
    \label{fig:hp-addition-neox}
\end{figure}

Additive steering only has two hyperparameters that were being optimized: the layer on which steering was done, and the beta value that determines the ``steering strength''. Figure \ref{fig:hp-addition-gptj} shows the performance of additive steering on the GPT-J model across all layers and beta values. Similarly to the results of conceptor-based steering, additive steering works best across layers 9-14 with peak performance always between layers 12-14. The best-performing beta values are 2.0, 3.0, and 4.0, although 2.0 is sufficient to reach peak performance for all tasks. 
Figure \ref{fig:hp-addition-neox} shows the performance of additive steering on the GPT-NeoX model across all layers and beta values. Similar to the best-performing conceptor-based steering hyperparameters, additive steering works best on layers 12-16. The optimal beta values are 1.5 and 2.0. 

